\documentclass[10pt,twocolumn,letterpaper]{article}
\usepackage{iccv}
\usepackage{times}
\usepackage{epsfig}
\usepackage{graphicx}
\usepackage{amsmath}
\usepackage{amssymb}

\usepackage{comment}
\usepackage{caption}
\usepackage{lipsum}
\usepackage{booktabs}
\usepackage{bbold}
\usepackage{amsthm,amsmath,amssymb}
\usepackage{mathrsfs}
\usepackage[noend]{algpseudocode}
\usepackage{algorithmicx,algorithm}
\usepackage{multirow}
\newcommand{\cmark}{\ding{51}}%
\usepackage{pifont}
\usepackage[mathscr]{euscript}
\usepackage{bbm}

\usepackage[breaklinks=true,bookmarks=false]{hyperref}

\iccvfinalcopy 

\def\iccvPaperID{2277} 

\ificcvfinal\fi

\begin{document}

\title{Point2Mask: Point-supervised Panoptic Segmentation via Optimal Transport}

\vspace{-3.0em}

\author{Wentong Li$^1$, \ \ \  Yuqian Yuan$^1$,  \ \ \ Song Wang$^1$, \\ Jianke Zhu$^{1}$\thanks{Corresponding author is Jianke Zhu.},  \ \ \  Jianshu Li$^2$, \ \ \ Jian Liu$^2$, \ \ \  Lei Zhang$^3$ \\
	$^1$Zhejiang University \ \ \ \ \ \  $^2$Ant Group \ \ \ \ \ \
	$^3$The HongKong Polytechnical University \\ 
}

\twocolumn[{%
\renewcommand\twocolumn[1][]{#1}%
\maketitle
\vspace{-2.8em}
\begin{center}
    \centering
    \includegraphics[width=0.98\textwidth]{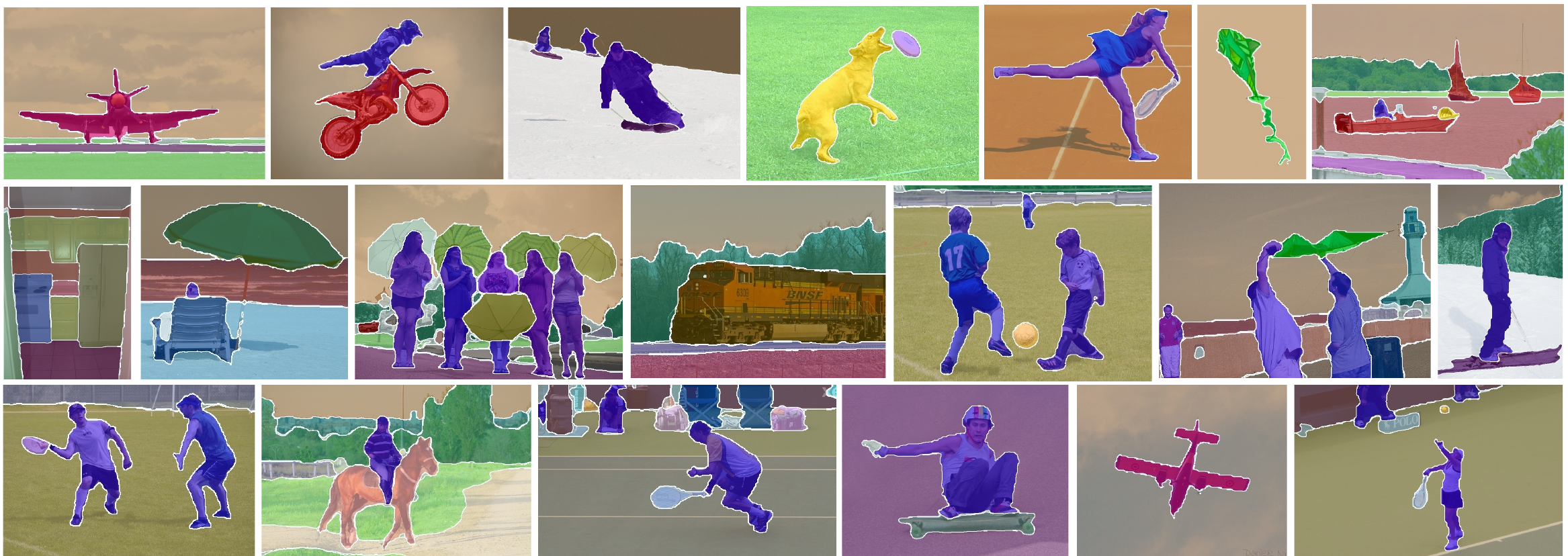}
\vspace{-0.5em}
\captionof{figure}{\textbf{Examples of pixel-wise mask predictions generated by Point2Mask} on COCO with ResNet-101. Only a \textit{single point} annotation per target is used as supervision during training to obtain these results.}
\label{fig:intro_vis}

\end{center}
}]

\ificcvfinal\thispagestyle{empty}\fi

\begin{abstract}
\vspace{-2mm}
Weakly-supervised image segmentation has recently attracted increasing research attentions, aiming to avoid the expensive pixel-wise labeling.
In this paper, we present an effective method, namely \textit{Point2Mask}, to achieve high-quality panoptic prediction using only a single random point annotation per target for training. 
Specifically, we formulate the panoptic pseudo-mask generation as an Optimal Transport (OT) problem, where each ground-truth ($gt$) point label and pixel sample are defined as the label supplier and consumer, respectively.  The transportation cost is calculated by the introduced task-oriented maps, which focus on the category-wise and instance-wise differences among the various thing and stuff targets. Furthermore, a centroid-based scheme is proposed to set the accurate unit number for each $gt$ point supplier. Hence, the pseudo-mask generation is converted into finding the optimal transport plan at a globally minimal transportation cost, which can be solved via the Sinkhorn-Knopp Iteration. Experimental results on Pascal VOC and COCO demonstrate the promising
performance of our proposed Point2Mask approach to point-supervised panoptic segmentation. 
Source code is available at: \href{https://github.com/LiWentomng/Point2Mask}{https://github.com/LiWentomng/Point2Mask}.

\end{abstract}

\section{Introduction}
Panoptic segmentation aims to obtain the pixel-wise labels of instance things and semantic stuff in the whole image, which plays an important role in applications such as autonomous driving, image editing and robotic manipulation. Although having achieved promising performance, most of the existing panoptic segmentation approaches~\cite{cvpr2022_panopticsegformer, nips2021_maskformer, zhang2021k, cheng2022masked, kirillov2019panoptic, yu2022cmt} are trained in a fully supervised manner, which heavily depend on the pixel-wise mask annotations, incurring expensive labeling costs.

To deal with this problem, weakly-supervised methods have recently attracted research attentions to obtain high-quality pixel-wise masks with label-efficient sparse annotations, such as bounding box~\cite{tian2021boxinst, li2022box,iccv2021discobox,li2022box2mask}, multiple points~\cite{li2022fully}, or the combination of them~\cite{cheng2022pointly, tang2022active}. 
Such methods make image segmentation
more accessible with lower annotation efforts for new categories or scene types. 
In this paper, we explore a simpler yet more efficient annotation form, \ie, \textit{a single random point} for each thing and stuff target, to achieve high-quality panoptic segmentation.
As discussed in~\cite{whatpoint_eccv2016}, the cost of point-level labels is only marginally above image-level ones~\footnote{On Pascal VOC~\cite{pascalvoc2010}, image labels cost around 20 sec./img, single point labels cost 22.1 sec./img, while full mask labels cost 239.7 sec./img.}.
Such a setting has been rarely studied due to the little available supervision information from a single point for pixel-wise mask prediction. Only one recent study~\cite{fan2022pointly} has attempted to build the minimum traversing distance between each pair of pixel sample and  ground-truth (denoted as $gt$) point label to determine the accurate pseudo mask label.

\begin{figure}[t]
\begin{center}
\includegraphics[width=0.999\linewidth]{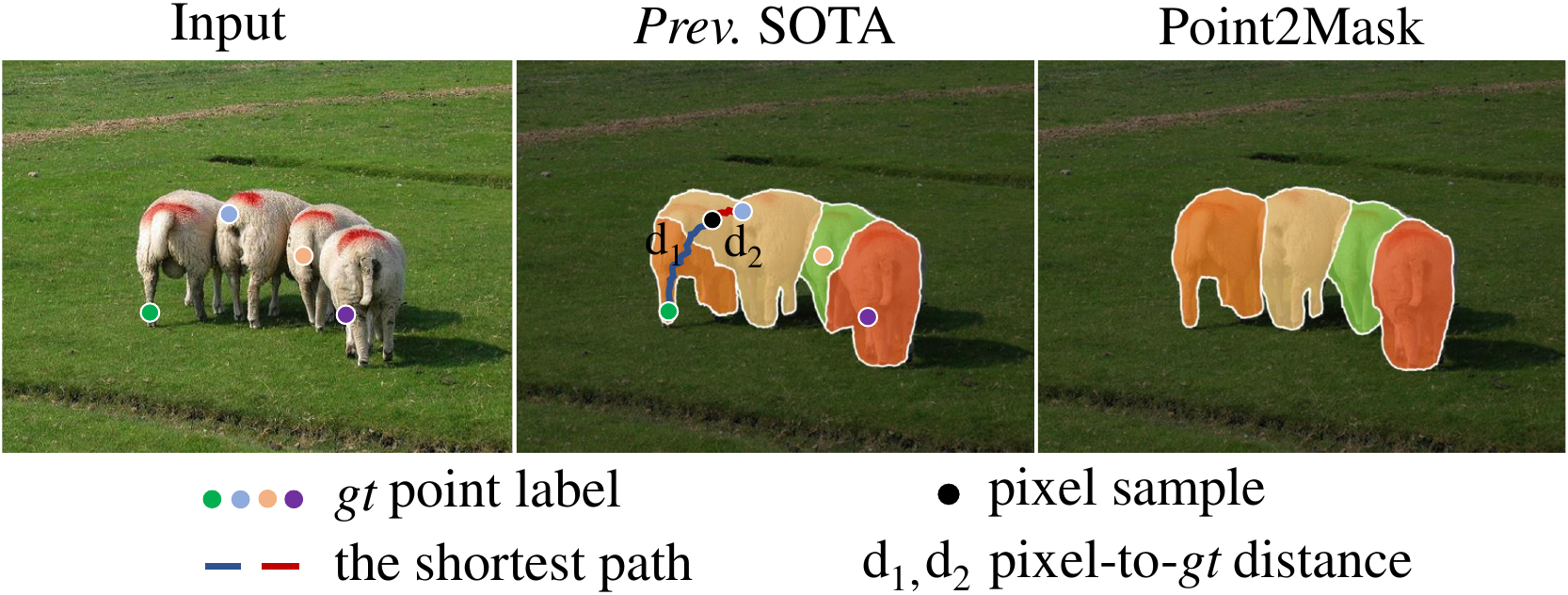} \end{center}
    \vspace{-1.0em}
    \caption{By taking an image with a single random $gt$ point label per target as the input, the method in~\cite{fan2022pointly} adopts the minimum distance for each pixel-$gt$ pair to determine the pseudo label, which cannot handle the ambiguous locations and heavily relies on the defined distance. For example, d$_2$ is shorter than d$_1$ for the current  pixel in black color, which results in wrong assignment.
    Our Point2Mask formulates this task as a global Optimal Transport problem, and obtains accurate pseudo-mask labels.}
\label{fig:intro_motivation}
\vspace{-4mm}
 \end{figure}

Unfortunately, it is sub-optimal to assign the pixel samples independently for each random $gt$ point label according to the  defined minimum distance.
As shown in Fig.~\ref{fig:intro_motivation},
the previous method \cite{fan2022pointly} heavily relies on the defined distance and lacks the global context in dealing with the ambiguous locations (\textit{i.e.}, the border pixels among different thing-based targets with the same category). 
The pixel-to-$gt$ assignment for ambiguous samples is non-trivial, which requires further information beyond the local view. To this end, we model this task from a global optimization perspective to determine the high-quality pixel sample partition for all $gt$ point labels within an image.

In this paper, we propose a novel single point-supervised panoptic segmentation method, dubbed as \textit{Point2Mask}, which formulates the pseudo-mask generation as an Optimal Transport (OT) problem. 
Specifically, we firstly define each $gt$ point label as a supplier who provides a certain number of labels, and regard each pixel sample as a consumer who needs one unit $gt$ label. To accurately define the transportation cost between each pixel-$gt$ pair, we introduce two types of task-oriented maps, including category-wise semantic map and instance-wise boundary map.
The former focuses on the semantic differences among the categories, while the later aims to discriminate the thing-based objects with accurate boundary. 
Furthermore, we propose an effective centroid-based scheme to set the accurate unit number for each $gt$ point supplier in the OT problem.

Under our proposed framework, the pseudo-mask generation is converted into finding the optimal transport plan at a globally minimal transportation cost, which can be efficiently solved via the Sinkhorn-Knopp Iteration~\cite{nips2013sinkhorndistances}.
By making use of the pseudo-mask labels, the panoptic segmentation sub-network is optimized in a fully-supervised manner. The proposed Point2Mask method is an end-to-end training framework, where only the fully-supervised sub-network is retained for inference. Extensive experiments are conducted on Pascal VOC~\cite{pascalvoc2010} and COCO~\cite{lin2014microsoft} benchmarks, and the promising qualitative and quantitative results demonstrate the effectiveness of our proposed approach. Notably, Point2Mask surpasses the state-of-the-art method~\cite{fan2022pointly} by 4.0\% PQ on Pascal VOC and 3.1\% PQ on COCO with the same ResNet-50 backbone~\cite{he2016deep}, and achieves comparable performance with the fully-supervised methods using the Swin-L backbone~\cite{liu2021swin}.  Some
qualitative results are shown in Fig.~\ref{fig:intro_vis}.

\section{Related Work}

\textbf{Fully-supervised Panoptic segmentation.} Image segmentation tackles the problem of grouping pixels. 
As the unified image segmentation task, panoptic segmentation~\cite{panopticfpn_cvpr2019} simultaneously incorporates semantic and instance segmentation, where each pixel is uniquely assigned with one of the stuff classes or one of the thing instances.  

To this end, some methods~\cite{panopticfpn_cvpr2019, cvpr2019upsnet, cvpr2020panopticdeeplab} have been proposed by dealing with things and stuff using  separate network branches within one model. Recently, some works~\cite{cvpr2022_panopticsegformer, nips2021_maskformer, maxdeeplab_cvpr2021, zhang2021k, cheng2022masked, maskdino} aim to unify the model for this task. DETR~\cite{eccv2020-detr} predicts the boxes for things and stuff categories with Transformer to perform panoptic segmentation. Mask2Former~\cite{cheng2022masked} further employs an additional pixel decoder to take into account of the high-resolution features and generates the mask predictions by the Transformer decoder with the masked-attention.
Despite being able to segment objects with accurate boundaries, these methods rely on the expensive and laborious pixel-wise mask annotations, which hinders them from dealing with new categories or scene types in real-world applications~\cite{whatpoint_eccv2016, tpami2023survey, iclrh2rbox}. 

\textbf{Weakly-supervised Panoptic Segmentation.}
Weakly supervised segmentation intends to alleviate the annotation burden in segmentation tasks by label-efficient sparse labels for training. According to different kinds of tasks, it ranges from semantic segmentation~\cite{TPAMI2021affinity,cvpr2022tree, iclr2021universal, cvpr2018normalized} to instance segmentation~\cite{cheng2022pointly, tian2021boxinst, iccv2021discobox, li2022box, li2022box2mask, ahn2019weakly} and to panoptic segmentation~\cite{fan2022pointly,cvpr2021_JTSM,li2022fully} tasks.
As for panoptic segmentation, Li \textit{et al.}~\cite{li2022fully} employed coarse polygons with multiple point annotations for each target to supervise the panoptic segmentation model.
Recently, Fan \textit{et al.}~\cite{fan2022pointly} adopted a simpler labeling form, \ie, a single point annotation, for each target in an image, and introduced the minimum traversing distance between each pixel sample and the target point label. In spite of its promising performance, it heavily relies on the defined distance, which cannot handle the ambiguous border locations with a local view. Thus, it is still challenging to obtain the accurate mask predictions for single point-supervised panoptic segmentation.

\textbf{Optimal Transport in Computer Vision.} The Optimal Transport (OT) is a classical optimization problem with a wide range of computer vision applications.
In the early years, the Wasserstein distance (WD), also known as the Earth Mover’s distance, was adopted to capture the structure of color distribution and texture spaces for image retrieval~\cite{rubner1998metric}. Recently, Chen \textit{et al.}~\cite{chen2020uniter} employed OT to explicitly encourage the fine-grained alignment between words and image regions for vision-and-language pre-training. 
Li \textit{et al.}~\cite{li2020enhanced} built an attention-aware transport distance in OT to measure the discriminant information from domain knowledge for unsupervised domain adaptation. 
To achieve high-quality label assignment, Ge \textit{et al.}~\cite{ge2021ota} formulated the label assignment in object detection as the problem of solving an OT plan.
In this work, we explore OT for point-supervised panoptic segmentation.

\begin{figure*}[t]
\begin{center}
\includegraphics[width=0.99\linewidth]{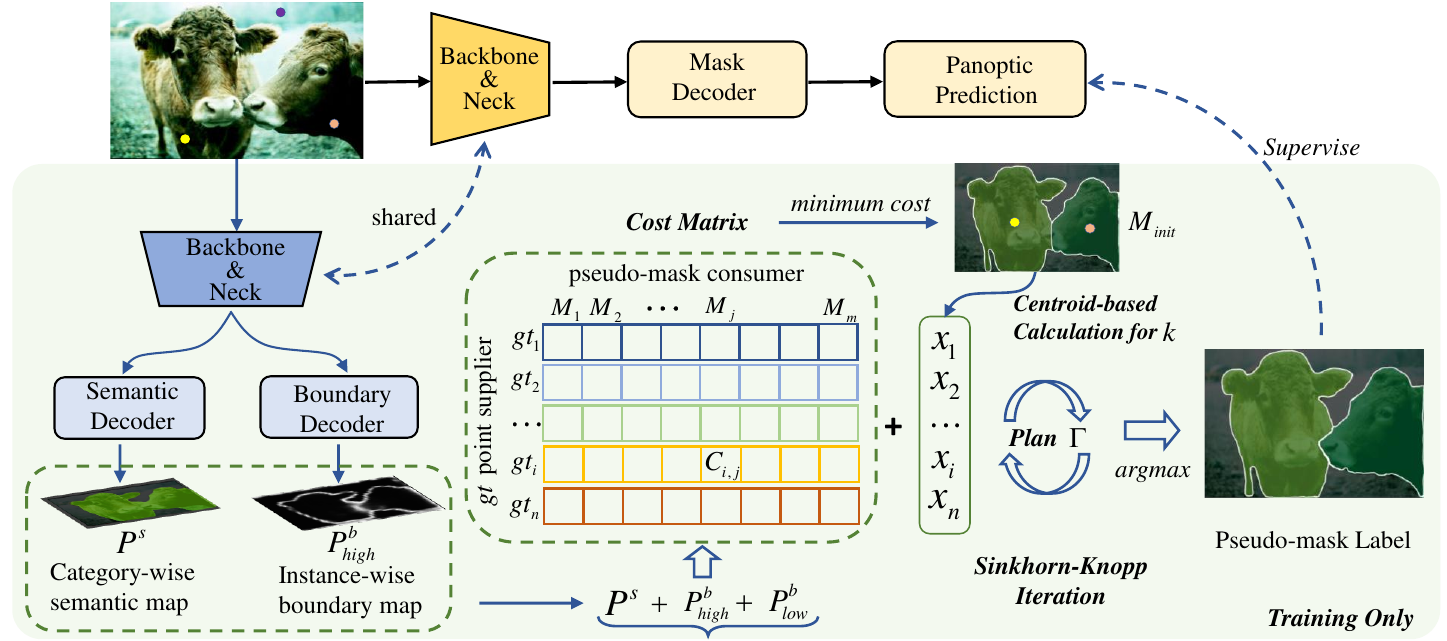} \end{center}
    \caption{Overview of Point2Mask. It consists of two branches, one branch for mask pseudo-label generation, and another for panoptic segmentation based on the generated pseudo-labels. The mask pseudo-label generation is formulated as the OT problem,
    where the cost matrix is defined based on the task-oriented maps. The $k$ unit number is calculated by the centriod-based scheme.
    The global optimal transportation plan ${\Gamma}$ can be solved by the Sinkhorn-Knopp Iteration to obtain the accurate pseudo-mask labels. Only panoptic segmentation branch is kept for inference.}
 \label{fig:overallnetwork}
 \end{figure*}

\section{Method}

\subsection{Overview of Point2Mask}
As illustrated in Fig.~\ref{fig:overallnetwork}, we leverage a unified framework, namely Point2Mask, for single point-supervised panoptic segmentation. 
It consists of two network branches. One branch generates the mask pseudo-labels, and the other focuses on the fully supervised learning using Panoptic SegFormer model~\cite{cvpr2022_panopticsegformer} based on the generated pseudo-labels.
The two branches share the basic backbone and neck network, which are trained in an end-to-end fashion.
The key of our proposed approach is how to model the process of mask pseudo-label generation as the global Optimal Transport (OT) problem, which aims to obtain the accurate pixel-wise pseudo-masks with only a single point label per target.

\subsection{Optimal Transport}

We first give a brief review of OT~\cite{rachev1985monge},  which aims to find a transportation plan $\Gamma$ minimizing the total cost of moving goods from one location to another. 
It is subject to certain constraints on the amount of goods to be transported and the cost of transportation. 

Given a set of $m$ suppliers, another set of $n$ consumers, and a cost function $c_{ij}$ that specifies the cost of transporting one unit of goods from the $i$-th supplier to the $j$-th consumer. The goal of OT is to find a transportation plan ${\Gamma} = \{ {\Gamma_{i,j}}\left| {i = 1,2, \cdots ,m,j = 1,2,} \right. \cdots ,n\}$ that minimizes the total cost of transporting all the goods from the suppliers to the consumers. Thus, the OT problem can be formulated as follows: 
\begin{equation}
\mathop {\min }\limits_{{\Gamma _{ij}} \in \Gamma } \;\;\sum\nolimits_{i,j}^{m,n} {{\Gamma _{ij}}{c_{ij}}},
\end{equation}
where $\Gamma_{ij} \ge 0$. The constraints to be satisfied are: the $i$-th supplier holds 
${x_i} = \sum\nolimits_{j = 1}^n {{\Gamma _{ij}}}$  units of goods, and the $j$-th consumer needs 
${y_j} = \sum\nolimits_{i =1}^m {{\Gamma _{ij}}}$ units goods. Meanwhile, the total amount of goods held by all suppliers are equal to the amount needed by all consumers, \ie, $\sum\nolimits_{i = 1}^m {{x_i}}  = \sum\nolimits_{j = 1}^n {{y_j}}$.
To efficiently tackle this problem, we adopt the Sinkhorn Iteration method~\cite{nips2013sinkhorndistances}. The details can be found in the Appendix.

\subsection{Pseudo-mask Generation by OT}

Given an input image $I^{H \times W  \times 3}$,  
supposing there are $m$ $gt$ point labels and $n$ pixel samples (\ie, $n=H\times W$), we view each $gt$ point label as a supplier who holds $k$ 
pixel samples (\ie, $x_i=k, i=1,2,..., m$).
Each pixel of $I$ is regarded as a consumer who needs one $gt$ point label (\ie, $y_j=1, j=1,2, ..., n$).
Given the defined cost $c_{ij}$ to transport one unit 
from the $i$-$th$ $gt$ point label to the $j$-th pixel,
the global OT plan $\Gamma \in \mathbb{R}^{m \times n}$ 
can be obtained by solving the OT problem via the Sinkhorn-Knopp Iteration~\cite{nips2013sinkhorndistances}. Once $\Gamma$ is obtained, the pseudo-mask label generation can be decoded by assigning the pixel samples to the suppliers who transport point $gt$ labels to them with the minimal transportation costs.

The pseudo-mask generation consists of task-oriented map generation, transportation cost definition  and centroid-based unit number calculation, which are introduced in details in the following subsections. The  completed procedure is summarized in Algorithm~\ref{OT_pseudo_mask}.

\subsubsection{Task-oriented Map Generation}
The task-oriented map includes the category-wise semantic map $P^{s}$ and instance-wise boundary map $P^{b}$. 
The former measures the semantic logit differences among the various categories. 
The latter discriminates the different thing-based targets under the same class from the accurate instance-level boundary.
Based on these maps, the distance of the adjacent pixels can be calculated to obtain each pixel-to-$gt$ cost $c_{ij}$.

\textbf{Category-wise Semantic Map.} 
An input image for panoptic segmentation task is composed of the stuff-based and thing-based targets. 
The semantic parsing is 
important to obtain category-wise logits.
As shown in Fig.~\ref{fig:overallnetwork},
we adopt the transformer decoder layers~\cite{cvpr2022_panopticsegformer} to construct the semantic decoder with a set of semantic query tokens, which is one-to-one match to the semantic categories.
The semantic logits $P^s$ with $N_c$ classes can be generated by multiplying the mask scores and the class probabilities together as in~\cite{fan2022pointly}.  The supervision information for category-wise semantic logits $P^s$ with the weak point labels is introduced in Sec.~\ref{weak loss} in detail.

\textbf{Instance-wise Boundary Map.} To discriminate the instances for thing-based targets, especially for the instances with the same category, we introduce the instance-wise boundary map $P^b$ for each target.

To generate the pure boundary, we suggest the high-level boundary $P_{high}^{b}$ that is learnt by the boundary decoder. In specific, we firstly sum the multi-level feature tokens from the Transformer-based neck in 2D spatial feature.
Then, two 1$\times$1 convolution layers interleaved by a ReLU activation are employed. The one-channel boundary map $P_{high}^{b}$ is obtained via the \texttt{sigmoid} function. For high-level boundary learning objective, we design an effective boundary loss function and explain it with details in Sec.~\ref{weak loss}.

Besides, we employ the Structured Edge (SE) detection method~\cite{iccv2013structured_edge} based on the original input image to capture the low-level contour $P_{low}^{b}$, which takes advantage of the inherent structure in edge patches to focus on the sparse object-level boundary map.

\subsubsection{Transportation Cost}

Based on the obtained task-oriented maps, the transportation cost can be calculated.

In our method, each map can be represented as an 8-connected planar graph $G(V,E)$, where each pixel is adjacent to eight neighbors. The vertex set $V$ consists of all pixels of the map, and the edge set $E$ is made of the edges between two adjacent vertices. Let the vertex $l$ and vertex $k$ be adjacent on the graph. Based on the $P^s$ and $P^b$ maps, the corresponding distance function  $d_{k,l}^s$ and $d_{k,l}^b$ can be defined as follows: 

\begin{equation}
\begin{split}
d_{k,l}^s &= \left| {P^s(k) - P^s(l)} \right|, \\
d_{k,l}^b &= \mathop {\max } \{ P^b(k),P^b(l)\},
\end{split}
\end{equation}
where ${P}(l)$, ${P}(k)$ are the map values of vertex $l$ and vertex $k$, respectively.
Once the edge length is obtained from the $P^s$ and $P^b$ maps, we define the transportation cost $c_{i,j}$ from the $i$-th pixel to the $j$-th $gt$ point label as the sum of the lengths of their connected edges along the shortest path $\mathbb{P}$:
\begin{equation}
{c_{i,j}} = \sum\limits_{(k,l) \in \mathbb{P}_{i,j}} ({d_{k,l}^s}  + \beta  {d_{k,l}^b}),
\label{edgedistance}
\end{equation}
where $\beta$ is the balanced weight. 
The shortest path $\mathbb{P}$ is implemented by the classical \texttt{Dijkstra} algorithm like~\cite{fan2022pointly}.

\begin{algorithm}[t]
\caption{Optimal Transport for Pseudo-mask Generation}
\label{alg:Algorithm1}
\hspace*{0.02in} {\bf Input:}\\
\hspace*{0.2in} $I^{H \times W \times 3}$ is an input image. \\
\hspace*{0.2in} $M^{H \times W \times 1}$ is the pseudo-mask label with ZerosInit. \\
\hspace*{0.2in} $\mathscr{P}$ is a set of $gt$ point labels. \\
\hspace*{0.2in} $T$ is the iteration number in Sinkhorn-Knopp Iter. \\
\hspace*{0.02in} {\bf Output:} \\
\hspace*{0.2in} $M$ is the assigned pseudo-mask label.
\begin{algorithmic}[1]

\State $m$ $\leftarrow$ $\left\vert \mathscr{P} \right\vert$, $n$ $\leftarrow$ $\left\vert M \right\vert$ \\
$P^{s}, P_{high}^{b}, P_{low}^{b} $ $\leftarrow$ Forward($I$, $\mathscr{P}$) \\
Compute pairwise pixel-to-$gt$ cost  $c_{ij}$.  \\
$x_i (i=1,2,...,m) \leftarrow$ Centriod-based $k$ calculation \\
$y_j (j=1,2,...,n) \leftarrow$  $\mathbb{1}$ \Comment{Init $y$ with ones} \\ 

$u^0, v^0\leftarrow$ $\mathbb{1}$ \Comment{Init $u$ and $v$ with ones}
\For{$t=0$ \textbf{to} $T$}: \\
\hspace*{0.2in} $u^{t+1},v^{t+1}\leftarrow$ SinkhornIter($c,u^t,v^t,x,y$) 
\EndFor \\
Compute optimal plan $\Gamma$. \\
Compute pseudo-mask label: $M= \text{argmax} (\Gamma). $
\State \Return $M$ 
\end{algorithmic}\label{OT_pseudo_mask}
\end{algorithm}

\subsubsection{Centroid-based Unit Number Calculation} \label{centriod-based}
\begin{figure}[t]
\begin{center}
\includegraphics[width=0.98\linewidth]{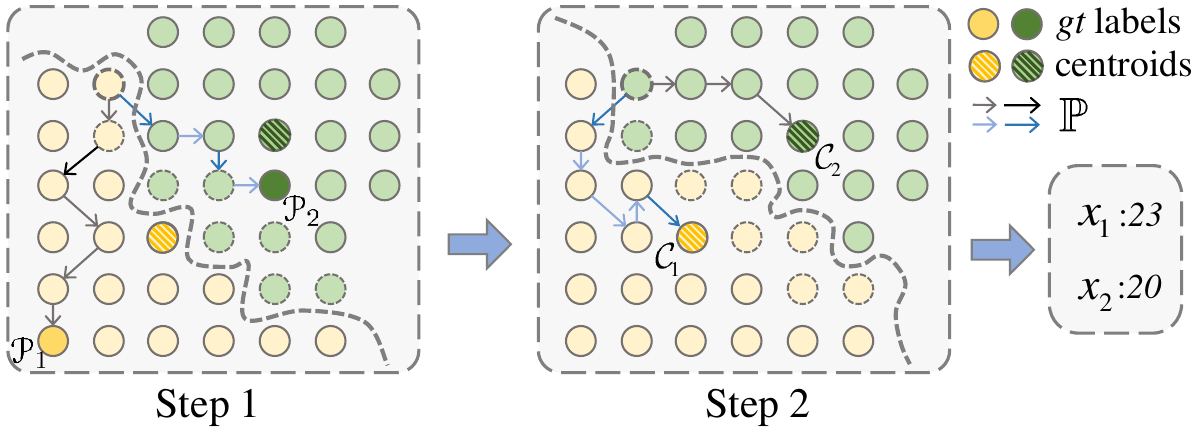} \end{center}
\vspace{-0.5em}
    \caption{The process of centroid-based $k$ calculation with two targets in an image.
    \textbf{Step 1:} The initial assignment (\ie, the pixels with yellow and green color divided by the middle curve line of dashes) with the minimal cost can be achieved based on the  $gt$ point labels $\mathscr{P}_1$ and  $\mathscr{P}_2$. 
    \textbf{Step2:} The centroids $\mathcal{C}_1$  and  $\mathcal{C}_2$ of each initially assigned mask are the  substitutions of $gt$ points, and the minimal cost can be re-calculated to achieve the refined assignment and determine the accurate unit number $k$  for each target.}
 \label{fig:centroids}
 \vspace{-2mm}
 \end{figure}

Each $gt$ point label $\mathscr{P}_i$ is regarded as the supplier in our proposed OT problem, which holds $x_i=k$ pixels of pseudo mask label $M$. To set the accurate number of $k$,  we introduce the centroid-based unit number calculation scheme that can be divided into two steps, as shown in Fig.~\ref{fig:centroids}.
 
Firstly, we obtain the pair-wise cost values along the shortest path $\mathbb{P}$ for each undetermined pixel to each $gt$ point label $\mathscr{P}_i$. 
The initial $gt$ point label assignment for each pixel can be achieved with its minimum cost among all $gt$ labels in the whole image.
Note that the $gt$ points are randomly labeled on each target in the image, which can be located at any position of the target to be segmented, such as the  corner or the edge. This cannot reflect the typical and accurate characteristics, especially for the border pixels between thing-based instances belonging to the same category.

Based on the initial $gt$ point label assignment, the initial mask label for each target can be obtained. 
We then calculate the corresponding centroid $\mathcal{C}_i$ of initial mask label as the substitution of $gt$ point label $\mathscr{P}_i$ for each target.
The pair-wise cost $c_{ij}$ for each pixel and $\mathcal{C}_i$ can be re-calculated along the corresponding shortest path. The $k$ unit number ($x_i$) is computed by counting the ones in $N_{ij}$ with the minimum cost values to each centriod $\mathcal{C}$,
which can be formulated as follows:
\begin{equation}
{x_i} = \sum\limits_j^n {{\textit{N}_{ij}}}, \ \ \ \ {\textit{N}_{ij}} = \left\{ \begin{array}{l}
1, \ \ \ \ \ \  {\rm{    }}\mathop {{\rm{argmin}}}\limits_i {c_{ij}} = i,\\
0,\ \ \ \ \ \ {\rm{     otherwise}}{\rm{.}}
\end{array} \right.
\end{equation}

The iterated calculation scheme can obtain a more accurate unit number $k$, and 
we leave the detailed performance analysis in Sec.~\ref{ablation} to examine the effectiveness of the proposed scheme.

\subsection{Learning and Inference}

\subsubsection{Weakly Supervised Learning}\label{weak loss}

In this section, we introduce the objective for category-wise semantic map $P^s$ and instance-wise boundary map $P^b$ in a weakly-supervised manner with only a single point label.

\textbf{Semantic Map Learning.} 
Like the weakly-supervised semantic  methods~\cite{cvpr2022tree,cvpr2018normalized},  we adopt the partial cross-entropy loss $\mathcal{L}_{partial}$, which is able to make full use of the available $gt$ point labels to achieve region supervised learning and generate sparse semantic map.

To obtain the accurate semantic logits for the unlabeled regions, we further take advantage of both local LAB affinity and long-range RGB affinity based on the input image. 
Local LAB affinity explores the color similarity in LAB color space with the local kernel, which is employed as the loss term $\mathcal{L}_{sem}^{LAB}$ as in~\cite{tian2021boxinst}. Long-range RGB affinity absorbs the pixel similarity in RGB space, which is implemented by the minimum spanning tree. As in~\cite{cvpr2022tree}, it is utilized as the loss term $\mathcal{L}_{sem}^{RGB}$. The objective for  semantic map learning is denoted as:
\begin{equation}
\mathcal{L}_{sem} = \mathcal{L}_{partial} + \alpha_1 \mathcal{L}_{sem}^{LAB}  +  \alpha_2 \mathcal{L}_{sem}^{RGB}.
\label{loss_sem}
\end{equation}
Please refer to the Appendix for the detailed formulation of these loss terms.

\textbf{High-level Boundary Map Learning.}
To encourage the boundary decoder to predict the high-level instance-wise boundary map $P_{high}^b$,
we suggest an effective loss function $\mathcal{L}_{bou}$ for panoptic segmentation task. 
In terms of the existence of a boundary between two adjacent pixels, we assume that their affinity is small as in~\cite{ahn2019weakly}.  Hence, we introduce the high-level affinity $\mathcal{A}$ representation.
For each pixel $p_k$ on $P_{high}^b$,  $p_l$ is one of its eight neighbors $\mathcal{N}_8$.
The $\mathcal{A}_{kl}$  can be represented as follows: 
\begin{equation}
{\mathcal{A}_{kl}} = 1 - \mathop {\max } {P_{high}^b}({p_k, p_l}).
\end{equation}
Then, we make full use of the mask affinity equivalence among the neighbor pixels based on the generated pseudo-mask $M$. 
The loss function $\mathcal{L}_{bou}$ can be defined as: 
\begin{footnotesize}
\begin{equation}
\begin{aligned}
{\mathcal{L}_{bou}} =  &- \sum\limits_{(k,l) \in M_{thing}^ + } {\frac{{\log {\mathcal{A}_{kl}}}}{{2\left| {M_{thing}^ + } \right|}}}  - \sum\limits_{(k,l) \in M_{stuff}^ + } {\frac{{\log {\mathcal{A}_{kl}}}}{{2\left| {M_{stuff}^ + } \right|}}} \\ & - \sum\limits_{(k,l) \in {M^ - }} {\frac{{\log (1 - {\mathcal{A}_{kl}})}}{{\left| {{M^ - }} \right|}}},
\end{aligned}
\end{equation}
\end{footnotesize}where $M^{+}_{thing}$ denotes that the pair of adjacent pixels $p_k$ and $p_l$ are inside the same thing-based pseudo mask. Similarly, $M^{+}_{stuff}$ represents that $p_k$ and $p_l$ are inside the same stuff-based pseudo mask. Instead, $M^{-}$ denotes that a pair of pixels are with different pseudo-mask labels. Driven by the ${\mathcal{L}_{bou}}$ term, we can learn the accurate high-level boundary. The Appendix show some visual examples for better illustration.

\subsubsection{Training and Inference}
\textbf{Loss Function.} Once the pseudo-masks are obtained, the panoptic segmentation sub-model is trained with these generated labels in a fully supervised manner.
We adopt Panoptic SegFormer~\cite{cvpr2022_panopticsegformer} as the panoptic sub-network.
The fully-supervised loss terms consist of the focal loss for classification prediction, the localization loss for box localization, and the dice loss on mask decoder for final panoptic segmentation,
respectively.  
For simplicity, we denote these losses to train the panoptic segmentation model as $\mathcal{L}_{full}$.
The total loss $\mathcal{L}_{total}$ can be formulated as follows:
\begin{equation}
\mathcal{L}_{total} = \mathcal{L}_{full} +  \mathcal{L}_{sem} + \mathcal{L}_{bou}.
\end{equation}

\begin{table*}[t]
\begin{center}
\setlength{\tabcolsep}{2.2mm}{
\begin{tabular}{llccccccc}
\hline
\multirow{2}{*}{Method} & \multirow{2}{*}{Backbone} & \multirow{2}{*}{Supervision}
 &\multicolumn{3}{c}{VOC 2012} & \multicolumn{3}{c}{VOC 2012 \textit{with} COCO} \\  
\cmidrule(r){4-6}
\cmidrule(r){7-9} 
& & & PQ & PQ$^{th}$ & PQ$^{st}$ & PQ & PQ$^{th}$ & PQ$^{st}$ \\

\hline\hline
Li~\textit{et al}.~\cite{li_eccv2018} & ResNet-101 & $\mathcal{M}$ & 62.7 & - & -& 63.1 & - & - \\

Panoptic FPN~\cite{panopticfpn_cvpr2019} &  ResNet-50 & $\mathcal{M}$ & 65.7 & 64.5 & 90.8 & - & - & -\\

Panoptic FCN~\cite{li2022fully} & ResNet-50 & $\mathcal{M}$  & 67.9  & 66.6 & \textbf{92.9} & \textbf{73.1} & \textbf{72.1} & \textbf{93.8}   \\

Panoptic SegFormer~\cite{cvpr2022_panopticsegformer} &  ResNet-50 & $\mathcal{M}$ & \textbf{67.9} & \textbf{66.6} & 92.7 & - & - & -\\

\hline

Li~\textit{et al.}~\cite{li_eccv2018} &  ResNet-101 & $\mathcal{B} + \mathcal{I}$ & 59.0 & - & - & 59.5 & - & -   \\

JTSM~\cite{cvpr2021_JTSM} & ResNet-18-WS~\cite{ResNet-18-ws-eccv2020} & $\mathcal{I}$  & 39.0 & 37.1 & 77.7 & - & - &- \\

PSPS~\cite{fan2022pointly} & ResNet-50 & $\mathcal{P}$ & 49.8 & 47.8 & 89.5   & - & - & -\\

Panoptic FCN~\cite{li2022fully} & ResNet-50 & $\mathcal{P}_{10}$  & 48.0 & 46.2 & 85.2 & 52.4 & 50.8 & 86.0 \\

\hline 
Point2Mask & ResNet-50& $\mathcal{P}$ & 53.8 & 51.9 & 90.5  & 60.7 & 59.1 & 91.8\\

Point2Mask & ResNet-101& $\mathcal{P}$ & 54.8 & 53.0 & 90.4 & 63.2 & 61.8 & 92.3\\

Point2Mask & Swin-L & $\mathcal{P}$ & \textbf{61.0} & \textbf{59.4} & \textbf{93.0} & \textbf{64.2} & \textbf{62.7} & \textbf{93.2}\\

\hline
\end{tabular}} 
\end{center}
\vspace{-1.2em}
\caption{Performance comparisons on Pascal VOC2012 $\texttt{val}$.
$\mathcal{M}$ denotes the pixel-wise mask annotations. $\mathcal{P}$ and $\mathcal{P}_{10}$ 
are point-level supervision with 1 and 10  points per target, respectively. $\mathcal{I}$ and $\mathcal{B}$ are the image-level and box-level supervisions (the same below).
 Besides, VOC 2012 \textit{with} COCO represents training and validation on VOC 2012 dataset with COCO pre-trained model.} \label{tab:voc}
\end{table*}

\textbf{Inference.} For the inference process of Point2Mask, 
only the panoptic segmentation model is maintained after training, which is the same as the original Panoptic SegFormer model~\cite{li2022fully}. The process of pseudo-mask generation with OT incurs about 25\% extra computational load in training, but it is totally cost-free during inference.

\begin{table*}[t]
\begin{center}
\setlength{\tabcolsep}{2.6mm}{
\begin{tabular}{llcccccccccc}
\hline
Method & Backbone & Supervision & PQ & PQ$^{th}$ & PQ$^{st}$ & SQ & RQ  \\

\hline\hline

AdaptIS~\cite{cvpr2019adaptis} &  ResNet-50 &  $\mathcal{M}$ & 35.9 & 40.3 & 29.3 & - & - \\

Panoptic FPN~\cite{panopticfpn_cvpr2019} &  ResNet-50 & $\mathcal{M}$ & 39.4 & 45.9 & 29.6 & 77.8 & 48.3 \\

Panoptic-DeepLab~\cite{cvpr2020panopticdeeplab} &  Xception-71~\cite{cvpr2017xception} & $\mathcal{M}$ & 39.7 & 43.9 & 33.2 & - & - \\

Panoptic FCN~\cite{li2022fully} & ResNet-50 & $\mathcal{M}$  &  43.6 & 49.3 & 35.0 & \textbf{80.6} & \textbf{52.6}  \\

Panoptic SegFormer~\cite{cvpr2022_panopticsegformer} &  ResNet-50 & $\mathcal{M}$  &  48.0 &52.3 & 41.5  & - & - \\

Mask2Former~\cite{cheng2022masked} & ResNet-50 & $\mathcal{M}$ &  \textbf{51.9} & \textbf{57.7} & \textbf{43.0} & - & -\\

\hline

JTSM~\cite{cvpr2021_JTSM} & ResNet-18-WS & $\mathcal{I}$ & 5.3 & 8.4 & 0.7 & 30.8  & 7.8  \\

PSPS~\cite{fan2022pointly} & ResNet-50 & $\mathcal{P}$  &  29.3 & 29.3 & 29.4  & - & - \\

Panoptic FCN~\cite{li2022fully} & ResNet-50 & $\mathcal{P}_{10}$  &  31.2 & 35.7  & 24.3 & - & -  \\

\hline 
Point2Mask & ResNet-50& $\mathcal{P}$ &  32.4 & 32.6 & 32.2 & 75.1 & 41.5  \\

Point2Mask & ResNet-101& $\mathcal{P}$ & 34.0 &  34.3 & 33.5 & 75.1 & 43.5   \\

Point2Mask & Swin-L & $\mathcal{P}$ & \textbf{37.0}  & \textbf{37.0} & \textbf{36.9}  & \textbf{75.8} & \textbf{47.2} \\

\hline
\end{tabular}}
\end{center}
\vspace{-1.0em}
\caption{Panoptic segmentation results on COCO \texttt{val2017}. Weakly and fully supervised methods are compared.} \label{tab:coco}
\vspace{-1.0em}
\end{table*}

\section{Experiments}
To evaluate our proposed approach, we conduct experiments on Pascal VOC~\cite{pascalvoc2010} and COCO~\cite{lin2014microsoft}. \textit{Only a single point label per target is used to train our method}, which is randomly sampled with the uniform distribution from the original pixel-wise mask annotations.

\subsection{Datasets}

\noindent  \textbf{Pascal VOC}~\cite{pascalvoc2010}. Pascal VOC consists of 20 ``thing'' and 1 ``stuff'' categories. 
It contains 10,582 images for model training and 1,449 validation images for evaluation~\cite{iccv2011_SBDdataset}.

\noindent  \textbf{COCO}~\cite{lin2014microsoft}. COCO has 80 ``thing"  and 53 ``stuff" categories, which is a challenging benchmark. Our models are trained on \texttt{train2017} (115K images), and evaluated on \texttt{val2017} (5K images).

\subsection{Implementation Details}
The models are trained with the AdamW optimizer~\cite{AdamW2017decoupled}.
We make use of the \texttt{mmdetection} toolbox~\cite{chen2019mmdetection}
and follow the commonly used training settings on each dataset. ResNet~\cite{he2016deep} and Swin-Transformer~\cite{liu2021swin} are employed as the backbones, which are pre-trained on ImageNet~\cite{IJCV2015imagenet}. 
On Pascal VOC, the initial learning rate is set to $10^{-4}$, and the weight decay is $0.1$ with eight images per mini-batch. The models are trained with $2 \times$ schedule at 24 epochs. On COCO, the initial learning rate is set to $2 \times 10^{-4}$, which is reduced by a factor of 10 at the 8-th epoch and 12-th epoch with 16 images per mini-batch. The models are trained with 15 epochs. 
The iteration number in Sinkhorn Iteration  for solving the defined OT problem is set to 80.
$\beta$ is 0.1 in Eq.~\ref{edgedistance}, and  $\alpha_1=\alpha_2=3.0$ in Eq.~\ref{loss_sem} in our implementation.
As in~\cite{li2022fully}, the number of query tokens for fully panoptic segmentation sub-model is set to 300.
The manifold projector proposed in~\cite{fan2022pointly} is employed to better stand for the instance-wise representation based on our baseline model. 
Unless specified, our centroid-based unit number calculation scheme is not iterated in the main experiments.
We report the standard evaluation metrics~\cite{panopticfpn_cvpr2019} of panoptic segmentation task, including panoptic quality (PQ), segmentation quality (SQ) and recognition quality (RQ).

\subsection{Main Results}
We compare our proposed Point2Mask method against state-of-the-art  weakly supervised  panoptic segmentation approaches. Moreover, the results of representative fully mask-supervised methods are reported for reference.

\textbf{Results on Pascal VOC.}  Table~\ref{tab:voc} reports the comparison results on Pascal VOC \texttt{val}.  
It can be clearly seen that Point2Mask with the ResNet-50 backbone outperforms the recent single point-supervised method PSPS~\cite{fan2022pointly} by absolute 4.0\% PQ (from 49.8\% to 53.8\%). The performance improvement mainly stems from the thing-based objects, from 47.8\% PQ$^{th}$ to 51.9\% PQ$^{th}$ (+4.1\% PQ$^{th}$), in contrast to the improvements on PQ$^{st}$ (89.5\% \textit{vs.} 90.3\%). It demonstrates the effectiveness of our presented pseudo-mask generation scheme by OT for thing-based instances. 
Our approach even outperforms Panoptic FCN~\cite{li2022fully} with 10 point labels by 5.8\% PQ (53.8\% \textit{vs.} 48.0\%). Moreover, our proposed method obtains 61.0\% PQ with Swin-L~\cite{liu2021swin} backbone, which  achieves comparable results against the fully supervised methods.
When the point-label COCO dataset is used for model pre-training, we achieve significant performance improvements, such as from 53.8\% PQ to 60.7\% PQ under the ResNet-50 backbone. With the  Swin-L backbone, Point2Mask obtains  64.2\% PQ,  surpassing the fully supervised method~\cite{li_eccv2018} by 1.1\% PQ.

\textbf{Results on COCO.} Table~\ref{tab:coco} gives the evaluation results comparing to the state-of-the-art (SOTA) methods on COCO. Our proposed Point2Mask method achieves 32.4\% PQ with single point supervision when ResNet-50 is employed as the backbone. 
It outperforms the previous SOTA method PSPS~\cite{fan2022pointly} by 3.1\% PQ, 3.3\% PQ$^{th}$ and 2.8\% PQ$^{st}$ under the same setting. 
Compared with  Panoptic FCN~\cite{li2022fully} with 10 point labels, our approach surpasses it by 1.2\% PQ (32.4\% \textit{vs.} 31.2\%). With Swin-L as the backbone, Point2Mask achieves 37.0\% PQ performance, which is comparable with some fully mask-supervised methods, including AdaptIS~\cite{cvpr2019adaptis}, Panoptic FPN~\cite{panopticfpn_cvpr2019} and Panoptic-DeepLab~\cite{cvpr2020panopticdeeplab} with ResNet-50 backbone.

\subsection{Ablation Studies} \label{ablation}
We analyze the design of each component in Point2Mask on Pascal VOC dataset.

\textbf{Different Task-oriented Maps.} 
We employ the category-wise semantic map $P^s$, low-level and  high-level boundary map $P_{low}^b$, $P_{high}^b$ to calculate the cost  for optimal transport.
Table~\ref{tab:taskmap} shows the evaluation results with different task-oriented maps. Our method achieves 50.6\% PQ using the $P^s$ map only, which focuses on the semantic logit differences among the categories. When $P_{low}^b$ and $P_{high}^b$ are employed separately, our method achieves 51.1\% PQ and 53.4\% PQ, respectively. More specifically, $P_{high}^b$ brings +2.9\% PQ gains driven by the designed boundary loss function $\mathcal{L}_{bou}$.
When all maps are adopted, Point2Mask achieves the best performance of 53.8\% PQ.

\textbf{Semantic Map Learning.}  Single point-supervised semantic parsing is the bedrock to obtain the panoptic segmentation results in our Point2Mask.
As shown in Table~\ref{tab:semanticmap}, when both  local LAB loss $\mathcal{L}_{sem}^{LAB}$ and  long-range RGB loss $\mathcal{L}_{sem}^{RGB}$ are adopted for the semantic map learning, the best 69.5\% mIoU and 53.8\% PQ are obtained comparing to each individual loss term.

\begin{table}[t]
\begin{center}
\setlength{\tabcolsep}{3.5mm}{
\begin{tabular}{ccccccc}
\hline
$P^s$ & $P_{low}^b$ & $P_{high}^b$ & PQ  & PQ$^{th}$ & PQ$^{st}$ \\
\hline\hline
\cmark &   &   &  50.6 & 48.7 & 90.1\\
\cmark & \cmark & &  51.1 & 49.1 & 90.3  \\
\cmark &   & \cmark & 53.4 & 51.6 & 90.3 \\
\cmark & \cmark & \cmark  &  \textbf{53.8} & \textbf{51.9} & \textbf{90.5} \\
\hline
\end{tabular}}
\end{center}
\vspace{-1.5em}
\caption{The impact of different task-oriented maps to calculate the pixel-to-$gt$ point label cost $c_{ij}$ in OT.}\label{tab:taskmap}
\vspace{-0.5em}
\end{table}

\begin{table}[t]
\begin{center}
\setlength{\tabcolsep}{1.5mm}{
\begin{tabular}{ccccccc}
\hline
$\mathcal{L}_{partial}$   & $\mathcal{L}_{sem}^{LAB}$  & $\mathcal{L}_{sem}^{RGB}$  & mIoU & PQ  & PQ$^{th}$ & PQ$^{st}$ \\
\hline\hline
\cmark &   &    &  61.6 & 40.4 & 38.1 & 86.1\\
\cmark & \cmark  &   &  69.0 & 51.2 & 49.3 & 90.0\\
\cmark &   &  \cmark  &  68.0 & 49.5 & 47.5 & 89.3 \\
\cmark & \cmark  & \cmark  & \textbf{69.5} & \textbf{53.8} & \textbf{51.9} & \textbf{90.5} \\ 
\hline
\end{tabular}}
\end{center}
\vspace{-1.5em}
\caption{Comparison of different weakly-supervised loss terms for category-wise semantic map learning.} 
\label{tab:semanticmap}
\vspace{-1.0em}
\end{table}

\begin{figure}[t]
\begin{center}
\includegraphics[width=0.99\linewidth]{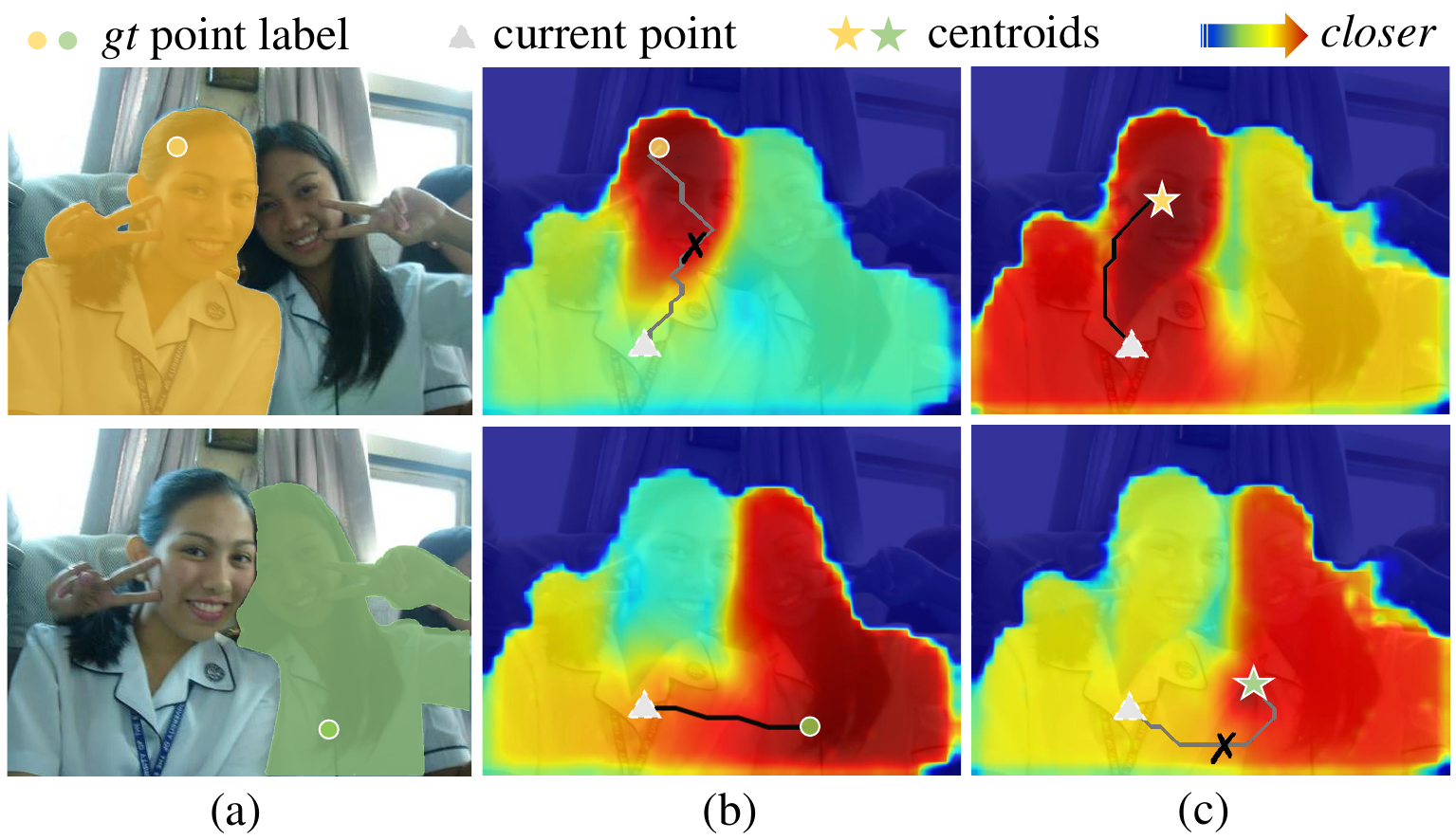} \end{center}
\vspace{-1.5em}
    \caption{ Visual comparisons on distance  heatmap with different calculation schemes of $k$. (a) shows the $gt$ point label and pixel-wise mask label. (b) indicates the  heatmap based on the Nearest $gt$ Point scheme. (c) is the heatmap based on our proposed Nearest Centroid scheme. The corresponding shortest paths are shown for better illustration.}
 \label{fig:heatmap}
 \vspace{-4.0mm}
 \end{figure}

\textbf{Different Unit Number Calculation Schemes.}
We explore three different schemes to calculate the unit number $k$ for $gt$ supplier, including ``Equal Division'', ``Nearest $gt$ Point'' and ``Nearest Centroid''. 
The Equal Division treats the mean value as $k$ for each $gt$ point supplier from all pixels.
The Nearest $gt$ Point indicates that the total number of pixels are with the nearest distances measured by the cost for each $gt$ point. For simplicity,  we denote the presented centroid-based unit number calculation scheme in Sec.~\ref{centriod-based} as the Nearest Centroid. Table~\ref{tab:k_cal} reports the comparison results.  Our Nearest Centroid scheme obtains the best performance with 53.8\% PQ, which outperforms Equal Division and Nearest $gt$ Point by 1.4\% PQ and 1.0\% PQ, respectively. 
Furthermore, we report the visual comparisons on distance heatmap, as shown in Fig.~\ref{fig:heatmap}. 
It can be clearly seen that the proposed Nearest Centroid scheme obtains the accurate unit number $k$ for each $gt$ point supplier.

In addition, as shown in Table~\ref{tab:iteracheme}, the Nearest Centroid scheme with more iterations (8 iterations) can bring a performance gain of +0.48\% PQ.  With 10 iterations, the model achieves the saturated performance with 54.07\% PQ.

\begin{table}
\begin{center}
\setlength{\tabcolsep}{4.5mm}{
\begin{tabular}{lccc}
\hline
 Scheme &   PQ  & PQ$^{th}$ & PQ$^{st}$ \\
\hline\hline
Equal Division     &  52.4 & 50.5 & 90.2\\
Nearest $gt$ Point  &  52.8 & 50.9 & 90.1 \\
Nearest Centriod  & \textbf{53.8} & \textbf{51.9} & \textbf{90.5}  \\
\hline
\end{tabular}}
\end{center}
\vspace{-1.0em}
\caption{Performance with different calculation schemes of $k$ for our defined OT problem in Point2Mask.}
\label{tab:k_cal}
\end{table}

\begin{table}
\begin{center}
\vspace{-1.0mm}

\begin{tabular}{cccccc}
\hline
Iterations  &  1  & 2  & 4 & 8 & 10 \\
\hline\hline
PQ  & 53.76  & 53.80  &  53.91 & \textbf{54.24}  & 54.07 \\
\hline
\end{tabular}
\end{center}
\vspace{-1.2em}
\caption{Performance with various iterations in centroid updating of the Nearest Centroid scheme.} \label{tab:iteracheme}
\vspace{-2.0mm}
\end{table}

\begin{table}
\begin{center}
\setlength{\tabcolsep}{1.8mm}{
\begin{tabular}{lccc}
\hline
Method  & PQ  & PQ$^{th}$ & PQ$^{st}$  \\
\hline\hline
Minimum Cost  &  51.9 &  50.1  &  90.2  \\
Optimal Transport  & \textbf{54.2}($\uparrow$2.3)  & \textbf{52.4}($\uparrow$2.3) & \textbf{90.3}($\uparrow$0.1)   \\
\hline
\end{tabular}} 
\end{center}
\vspace{-1.5em}
\caption{Comparisons between  Minimum Cost (MC)  and Optimal Transport (OT) based on the defined cost for pseudo-mask label generation.} \label{tab:otscheme}
\vspace{-4.5mm}
\end{table}

\textbf{Different Pseudo-mask Generation Methods.}
To examine the effectiveness of our proposed OT-based scheme, we study the different methods on pseudo-mask generation in Point2Mask.
Based on the presented cost on the task-oriented maps, we compared OT with the direct minimum cost (MC) method. Similar to~\cite{fan2022pointly}, MC assigns the $gt$ point label to each pixel with its corresponding minimum cost individually. 
Table~\ref{tab:otscheme} shows the comparison results. 
Point2Mask with our proposed OT method outperforms the MC scheme by +2.3\% PQ.
Specifically, the performance gains mainly stem from the thing-based targets (+2.3\% PQ$^{th}$ \textit{vs.} +0.1\% PQ$^{st}$). 
This is because it takes consideration of the global optimization in dealing with the
ambiguous locations, like the border pixels between different thing-based targets with the same category.

\section{Conclusion}
An effective single point-supervised panoptic segmentation approach, namely Point2Mask, was presented. The accurate pseudo-mask was obtained by finding the optimal transport plan at a globally minimal transportation cost, which was defined according to the task-oriented maps. Moreover, an effective centroid-based scheme was introduced to obtain the accurate unit number for each $gt$ point supplier. Extensive experiments were conducted on Pascal VOC and COCO benchmarks, validating the leading performance of the proposed Point2Mask over the previous state-of-the-arts on point-supervised panoptic segmentation.

\section*{Acknowledgments}
This work is supported by National Natural Science Foundation of China under Grants (61831015). Corresponding author is Jianke Zhu.

{\small
\bibliographystyle{ieee_fullname}
\bibliography{egbib}
}

\clearpage
\appendix
\section*{Appendix}

\setcounter{figure}{0}
\setcounter{table}{0}
\renewcommand{\thefigure}{A\arabic{figure}}
\renewcommand{\thetable}{A\arabic{table}}

\section{Sinkhorn Iteration}

The transport solver involves the resolution of a linear program in polynomial time. In our OT-based approach, the dimension of pixel samples can be as high as the square of hundreds. 
To efficiently tackle such a large-scale transport problem, we adopt the Sinkhorn Iteration method~\cite{nips2013sinkhorndistances,ge2021ota}, which computes the OT problem through the Sinkhorn’s matrix scaling algorithm. %

The Sinkhorn Iteration converts the OT optimization target
into a non-linear but convex form with an entropic regularization term $R$, which can be formulated as below:
\begin{equation}
\mathop {\min }\limits_{{\Gamma _{ij}} \in \Gamma } \ \ \sum\limits_{i,j = 1}^{m,n} {{\Gamma_{ij}}{c_{ij}}}  + \lambda R({\Gamma_{ij}}),
\end{equation}
where $R({\Gamma_{ij}}) = {\Gamma _{ij}}(\log {\Gamma _{ij}} - 1)$, and $\lambda$ is a regularization coefficient. According to the Sinkhorn-Knopp Iteration method~\cite{nips2013sinkhorndistances, sinkhorn1967}, $v_i$ and $u_j$ are introduced for updating the solution: 
\begin{equation}
u_j^{t + 1} = \frac{{{y_j}}}{{\sum\limits_i {{K_{ij}}} v_i^t}}, \ \ \ v_i^{t + 1} = \frac{{{x_i}}}{{\sum\limits_j {{K_{ij}}} u_j^{t + 1}}},
\end{equation}
where ${K_{ij}} = e^{( - {{{c_{ij}}}} \mathord{\left/
 {\vphantom {{{c_{ij}}} \lambda }} \right.
 \kern-\nulldelimiterspace} \lambda )}$. After performing the iteration for $T$ times, the optimal plan $\Gamma$ can be obtained as:
\begin{equation}
\Gamma  = \mathbf{diag}(u)K\mathbf{diag}(v).
\end{equation}

\section{Semantic Map Learning}

The local LAB affinity and the long-range RGB affinity are integrated to generate the accurate semantic map $P^s$ for the unlabeled regions. In the following, we introduce the two loss terms in detail.

\textbf{Local LAB Loss.}  As in~\cite{tian2021boxinst}, the local LAB loss $\mathcal{L}_{sem}^{LAB}$ explores the color similarity $\mathcal{S}_{LAB}$ in LAB color space of the input image with the local kernel. $\mathcal{S}_{LAB}$ is defined as:
\begin{equation}
{\mathcal{S}_{LAB}} = \mathop {\mathcal{S}({r_{i}},{r_{j}})}\limits_{j \in {\mathcal{N}_8}(i)}  = \exp\left  ( - \frac{{\left\| {{r_{i}} - {r_{j}}} \right\|}}{\theta_1 }\right),
\end{equation}
where $r_{i}$ is the LAB color value of pixel $i$ and $\mathcal{N}_8(i)$ denotes its eight local neighbors. 
${\theta}_1$ is the  constant parameter.  
The $\mathcal{L}_{sem}^{LAB}$ loss term is formulated as follows: 
\begin{equation}
{\mathcal{L}_{sem}^{LAB}} =  - \frac{1}{z_1}\sum\limits_{i = 1}^{n} {\sum\limits_{j \in {\mathcal{N}_8}(i)} {{\mathbbm{1}_{\{ {{\mathcal{S}}_{i,j}^{LAB}} \ge \tau \} }}{\log {P^s_i}^T P^s_j}}},
\end{equation}
where $z_1 = \sum\nolimits_{i = 1}^n {\sum\nolimits_{j \in {\mathcal{N}_8}(i)} {{\mathbbm{1}_{\{ {{\mathcal{S}}_{i,j}^{LAB}} \ge \tau \} }}} }$. ${{\mathbbm{1}_{\{ {{\mathcal{S}}_{i,j}^{LAB}} \ge \tau \} }}}$ is the indicator function, being 1 if $\mathcal{S}_{i,j}^{LAB} \ge \tau$ and 0 otherwise. 
As in~\cite{tian2021boxinst}, $\tau$ is set to $0.3$ and ${\theta}_1$ is set to 2 by default.

\textbf{Long-range RGB Loss.} Similar to~\cite{cvpr2022tree}, the long-range RGB loss  $\mathcal{L}_{sem}^{RGB}$ absorbs the global pixel affinity in RGB space. 
 Each pixel in the input image can be constructed by the global RGB pixel similarity $\mathcal{S}_{RGB}$ through the minimum spanning tree (MST) algorithm. The pixel similarity $\mathcal{S}_{RGB}$ in each tree-connected edge $\mathbb{E}$ is defined as follows:
\begin{equation}
{{\cal S}_{RGB}} = \mathop {{\cal S}({r_i},{r_j})}\limits_{(l,k) \in \mathbb{E}(i,j)}  = \exp \left( - \frac{{\sum {{{\left\| {{r_l} - {r_k}} \right\|}^2}} }}{{{\theta _2}}}\right),
\end{equation}
where $r_i$ is the RGB pixel value of pixel $i$. $l$ and $k$ are the adjacent pixels in the tree-connected edge $\mathbb{E}_{i,j}$. Like ${\theta}_1$,  ${\theta}_2$ is a constant value, which is set to 0.02 by default.
The $\mathcal{L}^{sem}_{RGB}$ loss term is defined as: 
\begin{equation}
{\mathcal{L}}_{sem}^{RGB} =  - \frac{1}{n}\sum\limits_{i = 1}^n {\left| {P_i^s - \frac{1}{{{z_2}}}\sum\limits_{\forall j \in \Omega } {{\mathcal{S}}_{i,j}^{RGB}P_j^s} } \right|},
\end{equation}
where ${z_2} = \sum\nolimits_{j} {\mathcal{S}_{i,j}^{RGB}}$,  $\Omega$ denotes the set of pixels in $P^s$.

\begin{table*}[t]
\begin{center}
\setlength{\tabcolsep}{2.2mm}{
\begin{tabular}{llccccccc}
\hline
\multirow{2}{*}{Method} & \multirow{2}{*}{Backbone} & \multirow{2}{*}{Supervision}
 &\multicolumn{3}{c}{VOC 2012} & \multicolumn{3}{c}{COCO} \\  
\cmidrule(r){4-6}
\cmidrule(r){7-9} 
& & & PQ & PQ$^{th}$ & PQ$^{st}$ & PQ & PQ$^{th}$ & PQ$^{st}$ \\

\hline\hline

Panoptic FPN~\cite{panopticfpn_cvpr2019} &  ResNet-50 & $\mathcal{M}$ & 65.7 & 64.5 & 90.8 & 41.5  & 48.3  &  31.2 \\

Panoptic FCN~\cite{li2022fully} & ResNet-50 & $\mathcal{M}$  & 67.9  & 66.6 & \textbf{92.9} &  43.6 &  49.3 &  35.0  \\

Panoptic SegFormer~\cite{cvpr2022_panopticsegformer} &  ResNet-50 & $\mathcal{M}$ & \textbf{67.9} & \textbf{66.6} & 92.7 &  \textbf{48.0} & \textbf{52.3}  & \textbf{41.5}  \\

\hline

PSPS~\cite{fan2022pointly} & ResNet-50 & $\mathcal{P}$ & 49.8 & 47.8 & 89.5 & 29.3 &  29.3 &  29.4 \\
Point2Mask (Ours) & ResNet-50 & $\mathcal{P}$ & 54.2 & 52.4 & 90.3  & 32.4 & 32.6 & 32.2 \\

\hline

Panoptic FCN~\cite{li2022fully} & ResNet-50 & $\mathcal{P}_{10}$  & 48.0 & 46.2 & 85.2 & 31.2 & 35.7 & 24.3  \\

PSPS~\cite{fan2022pointly} & ResNet-50 & $\mathcal{P}_{10}$ & 56.6 & 54.8 & 91.4 & 33.1  & 33.6  & 32.2 \\
 
Point2Mask (Ours) & ResNet-50 & $\mathcal{P}_{10}$ & 59.1 & 57.5 &  91.8 & 35.2  & 36.1 & 34.0  \\

Point2Mask (Ours) & ResNet-101 & $\mathcal{P}_{10}$ & \textbf{60.2}  & \textbf{58.6} & \textbf{92.1} & \textbf{36.7}  & \textbf{37.3}  &  \textbf{35.7}\\

\hline
\end{tabular}} 
\end{center}
\vspace{-1.5em}
\caption{Performance comparison on Pascal VOC $\texttt{val}$ and COCO $\texttt{val2017}$. 
$\mathcal{M}$ is pixel-wise mask label. $\mathcal{P}$ and $\mathcal{P}_{10}$ denote 1 and 10 point labels per target, respectively.  The results with $\mathcal{M}$ and  $\mathcal{P}$ supervision are listed as reference to illustrate the performance with 10 point labels.} \label{tab:tenpoints}
\vspace{-3.5mm}
\end{table*}

\begin{table}
\begin{center}
\setlength{\tabcolsep}{4.1mm}{
\begin{tabular}{cccc}
\hline
 Iter. Num. &   PQ  & PQ$^{th}$ & PQ$^{st}$ \\
\hline\hline
40     & 53.0  & 51.2 & 90.1 \\
60  &  53.5 & 51.7 & 90.1 \\
80  & \textbf{53.8} & \textbf{51.9}  & \textbf{90.5}  \\
100  & 52.7 & 50.8 & 90.1  \\
120  & 52.2 & 50.3 &  90.2  \\
\hline
\end{tabular}}
\end{center}
\vspace{-1.5em}
\caption{The results with different number of iterations in the Sinkhorn Iteration.}
\label{tab:iter_num}
\vspace{-3.0mm}
\end{table}

\section{Additional Results}
\subsection{Performance on Multiple Point Labels}
To further investigate the effectiveness of our approach with multiple point labels,  we conduct the experiments with ten-points annotation per target. 
The results of 
fully mask-supervised and single point-supervised methods are also listed as reference.
As shown in Table~\ref{tab:tenpoints}, 
we compare Point2Mask with the state-of-the-art methods, including Panoptic FCN~\cite{li2022fully} and  PSPS~\cite{fan2022pointly} with ten-points labels on  Pascal VOC and COCO datasets.  
With ResNet-50 backbone, Point2Mask outperforms Panoptic FCN~\cite{li2022fully} by 11.1\% PQ (59.1\% \textit{vs.} 48.0\%) on Pascal VOC and 4.0\% PQ (31.2\% \textit{vs.} 35.2\%) on COCO. Compared with PSPS~\cite{fan2022pointly}, Point2Mask  surpasses PSPS~\cite{fan2022pointly} by 2.5\% PQ  and 2.1\% PQ on Pascal VOC and COCO, respectively.
Furthermore, Point2Mask achieves more competitive performance with 60.2\% PQ on Pascal VOC and 36.7\% PQ on COCO using ResNet-101 backbone.

\begin{table}
\begin{center}
\setlength{\tabcolsep}{5.0mm}{
\begin{tabular}{cccc}
\hline
$\beta$ & PQ  & PQ$^{th}$ & PQ$^{st}$ \\
\hline\hline
1.0 & 52.3  & 50.4  & 90.2 \\
0.5  & 52.4 &  50.5 & 90.2  \\
0.2  & 52.8 & 50.9 &  90.3 \\
0.1  & \textbf{53.8} & \textbf{51.9} &  \textbf{90.5} \\
0.05  & 53.1 & 51.2 &  90.1 \\
0.01  & 51.9 & 50.0 & 89.6 \\\hline
\end{tabular}}
\end{center}
\vspace{-1.4em}
\caption{Results with different values of $\beta$ in Eq.~\ref{edgedistance}.}
\label{tab:balance_eq3}
\vspace{-6.0mm}
\end{table}

\subsection{Hyper-parameter Selection in OT}
We perform the following experiments to examine the impact of hyper-parameters in our OT-based method.

\textbf{Different Number of Sinkhorn Iterations.} 
We perform Sinkhorn Iteration with different number of iterations to solve the OT problem.
Table~\ref{tab:iter_num} reports the panoptic segmentation results. When the iteration number is set to 80, Point2Mask achieves the best performance with 53.8\% PQ.

\begin{figure}[t]
\begin{center}
\includegraphics[width=0.999\linewidth]{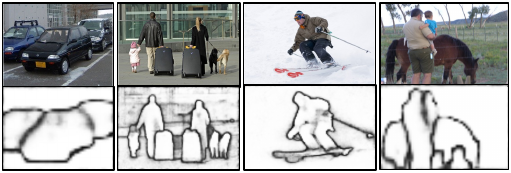} \end{center}
    \vspace{-1.5em}    
    \caption{Visual examples of high-level boundary map. The  accurate boundary for thing-based objects can be learnt.}  
 \label{fig:boundary}
 \vspace{-5mm}
 \end{figure}

\textbf{Impact of $\beta$.}  In our paper, $\beta$ in Eq.~\ref{edgedistance} indicates the importance of boundary map $P^b$ to calculate the pixel-to-$gt$ cost $c_{i,j}$.  Table~\ref{tab:balance_eq3} shows the results with different values of $\beta$. When $\beta=0.1$, Point2Mask obtains the best performance. 
This indicates that the cost from instance-wise boundary map $P^b$ plays a complementary role to the main cost term based on $P^s$. Furthermore, the visual examples of learnt high-level boundary $P^b_{high}$ are shown in Fig.~\ref{fig:boundary}.

\vspace{-0.6mm}
 
\subsection{More Visualization Results}
To further illustrate the performance of our single point-supervised approach, we give more visualization results. 

\begin{figure*}[t]
\begin{center}
\includegraphics[width=0.95\linewidth]{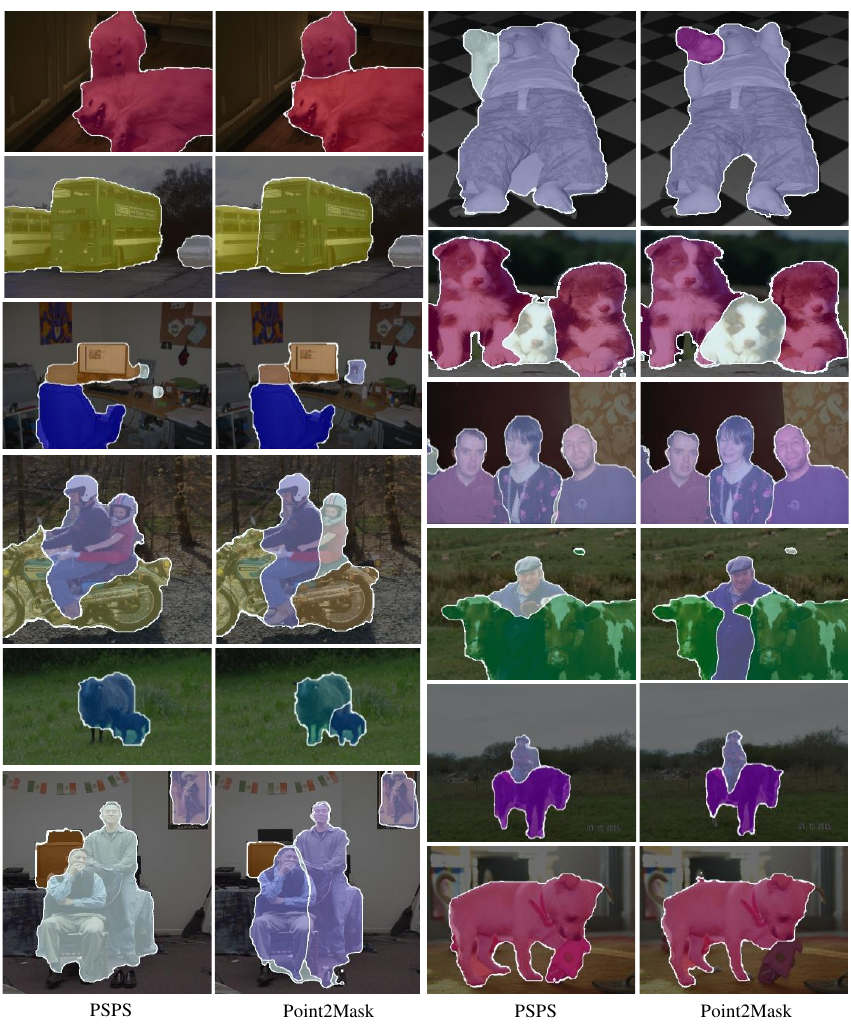} \end{center}
\vspace{-3.0mm}
    \caption{Qualitative comparisons on  Pascal VOC.
    The left two columns show that Point2Mask can precisely discriminate the nearby instances of the same category. The right two columns indicate that Point2Mask can obtain more fine-grained boundaries.}
 \label{fig:visual_compare}
 \vspace{-3.0mm}
 \end{figure*}

\begin{figure*}[t]
\begin{center}
\includegraphics[width=0.999\linewidth]{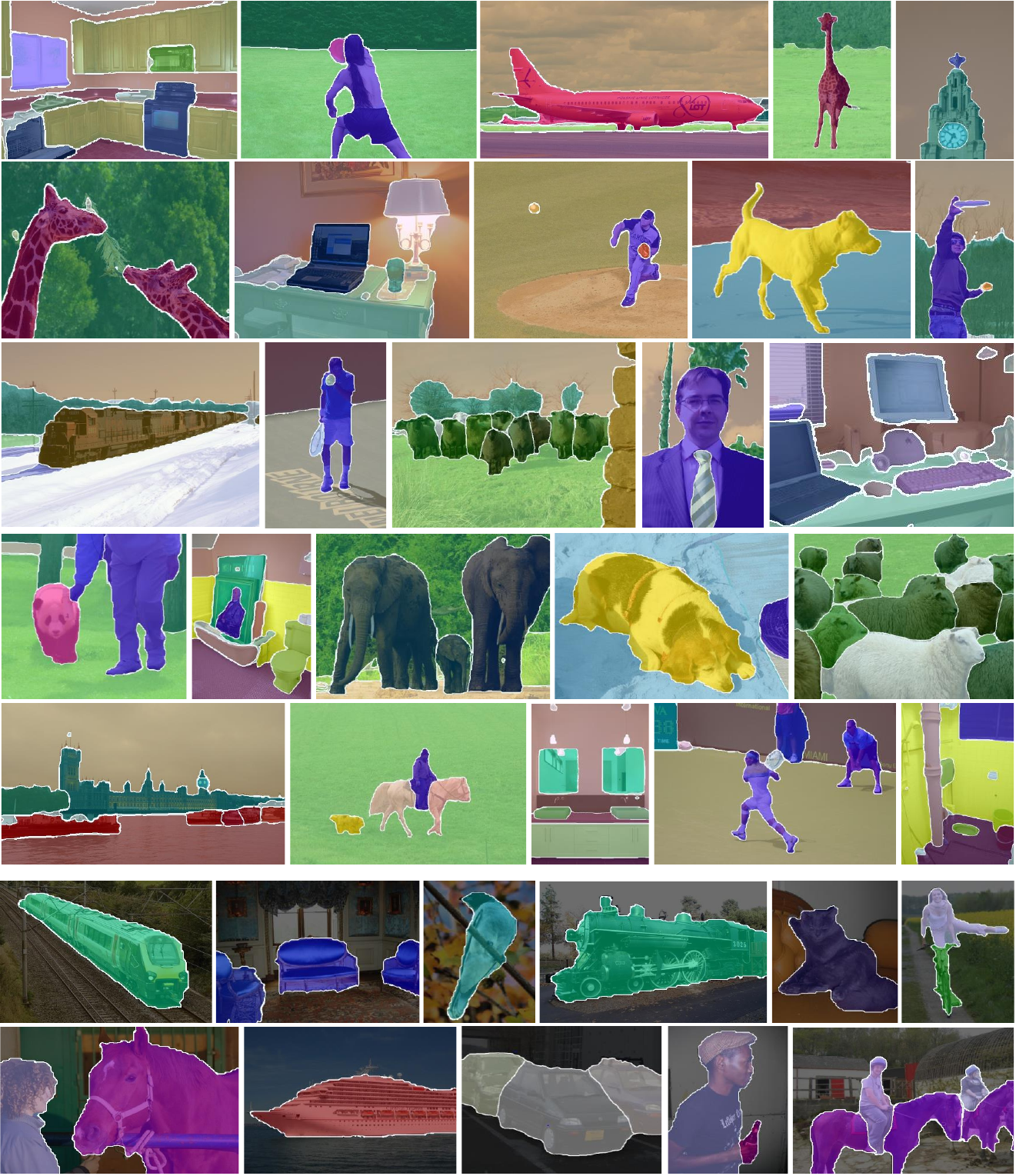} \end{center}
    \caption{Visual examples of panoptic segmentation by our Point2Mask with single point label per target on COCO and Pascal VOC datasets.}
 \label{fig:visual_coco}
 \end{figure*}

Fig.~\ref{fig:visual_compare} shows the qualitative comparison with the state-of-the-art method PSPS~\cite{fan2022pointly}.
It can be seen that our proposed Point2Mask approach is able to find the ambiguous locations of nearby instances precisely. This demonstrates that our OT-based approach can discriminate the thing-based targets with the accurate boundaries. In addition, Fig.~\ref{fig:visual_coco} provides the panoptic segmentation results of Point2Mask on general COCO and Pascal VOC datasets.

\section{Discussion}

\textbf{Differences against the existing works.} Like previous weakly-supervised methods~\cite{fan2022pointly, tian2021boxinst, li2022box2mask, li2022box}, our method aims to achieve high-quality segmentation with the label-efficient sparse labels, which is different from the existing promptable segmentation model~\cite{kirillov2023segment} with a large amount of data and  the corresponding mask labels.

We adopt the same base architecture as PSPS~\cite{fan2022pointly}, \textit{i.e.},  generating pseudo labels firstly and then training the panoptic segmentation branch. 
To generate the panoptic pseudo labels, both our method and PSPS~\cite{fan2022pointly} employ the category-wise and instance-wise representations.  
For category-wise representation, we firstly employ the local LAB and long-range RGB pixel similarities (Sec.3.4.1), instead of the local LAB semantic parsing only as in~\cite{fan2022pointly}. Secondly, for instance-wise representation, we adopt the boundary map and define different distance functions. 
Compared with the high-level manifold cues in~\cite{fan2022pointly}, the boundary map is more suitable for the shortest path-based implementation to calculate the instance-wise differences. 
More importantly, \textit{the key difference lies in the presented OT formulation for global assignment to generate more accurate mask labels}.
 
\textbf{Limitations.} For the dense objects with the same categories, such as in autonomous driving  and remote sensing scenarios, the proposed method may not perform well with the supervision of only a single point label. Better performance can be obtained by adopting the more powerful segmentation network, like Mask2Former~\cite{cheng2022masked} and MaskDINO~\cite{maskdino}, into our method.

\end{document}